\documentclass[letterpaper, 10 pt, conference]{ieeeconf}  

\IEEEoverridecommandlockouts                              

\title{\LARGE \bf
Feminist Perspective on Robot Learning Processes}

\author{Juana Valeria Hurtado$^{*1}$ and Valentina Mej\'ia$^{*}$ 
\thanks{* These authors have contributed equally to this work}
\thanks{$^{1}$ Department of Computer Science, University of Freiburg, Freiburg im Breisgau, Germany}%
}

\begin{document}

\maketitle
\thispagestyle{empty}
\pagestyle{empty}

\begin{abstract}
As different research works report and daily life experiences confirm, learning models can result in biased outcomes. The biased learned models usually replicate historical discrimination in society and typically negatively affect the less represented identities. Robots are equipped with these models that allow them to operate, performing tasks more complex every day. The learning process consists of different stages depending on human judgments. Moreover, the resulting learned models for robot decisions rely on recorded labeled data or demonstrations. Therefore, the robot learning process is susceptible to bias linked to human behavior in society. This imposes a potential danger, especially when robots operate around humans and the learning process can reflect the social unfairness present today. Different feminist proposals study social inequality and provide essential perspectives towards removing bias in various fields. What is more, feminism allowed and still allows to reconfigure numerous social dynamics and stereotypes advocating for equality across people through their diversity. Consequently, we provide a feminist perspective on the robot learning process in this work. We base our discussion on intersectional feminism, community feminism, decolonial feminism, and pedagogy perspectives, and we frame our work in a feminist robotics approach. In this paper, we present an initial discussion to emphasize the relevance of feminist perspectives to explore, foresee, en eventually correct the biased robot decisions.
\end{abstract}

\section{INTRODUCTION}
In the history of robotics, an important goal has been to equip robots with the ability to make complex decisions, and in some situations, with better effectiveness and precision than humans. These skills have already been demonstrated in various applications such as space exploration, medical applications, scene understanding, transport of materials, robots in the industry, and manufacturing and assembly, among many others 
For robots to correctly operate and achieve a required task, they need to sense their surroundings, identify and understand the elements and actions occurring around them, and interact with the world to perform a specific task. Additionally, in social environments, besides completing the task, we also expect that robots behave within certain social parameters similar to the decisions humans make, so the shared space is safe and comfortable for us. Machine learning algorithms have played an essential role in developing robots, making learning processes widely used in the field.  

With the increased precision and effectiveness a robot can execute a task, at some point, it is given higher autonomy to make decisions executing more complex tasks nearer to humans. This allows robots to plan a navigation route, manipulate objects, compute object recognition under challenging situations, interact with people, among many others. However, granting autonomy of decisions to robots based solely on technical advances without taking into account the social consequences and analyzing the development process is potentially dangerous. This risk is particularly concerning in a society still affected by the historical discrimination and oppression of diverse populations. Indeed, recently it has been demonstrated that algorithms and robots can reflect and even amplify these social problems, particularly affecting identities and communities that are usually the least represented \cite{bolukbasi2016man, caliskan2017semantics, buolamwini2018gender, fuchs2018dangers, west2020redistribution,garcia2016racist}.

Since there is a learning process for robots to operate and these robots interact with diverse groups of people, it is important then to question: Which learning techniques are utilized? What information is used in these methodologies and who is involved in the development process? Robotics is thought of as a human advance. Why not consider robot learning as an integral process towards the emancipation of human biases so that technology can disrupt traditional inequities? With these questions in mind, we argue that a feminist perspective in the robot learning process can first facilitate the identification of possible discriminating behaviors. Second, foresee biased outcomes that affect certain people by understanding how the learning process is susceptible to induce biased results. Third, promote changes in society by a more conscious development of robots that influence adequate behaviours.

In this paper, we specifically consider intersectional feminism , community feminism, decolonial feminism  and pedagogy perspectives  to provide an initial analysis to the learning process. We also aim to promote the discussion about robot behavior and feminism, by highlighting the relevance of a critical analysis of their interaction with humans. In this paper we do not analyze the gender that is attributed to robots. Instead, we explore how the behavior that robots learn can produce disproportionate effects on identities and communities. Additionally, we discuss how including a feminist perspective in the learning process can be reflected in the human-robot interaction. Therefore, to facilitate the analysis, we assume that robots are not perceived with a specific gender.

Our work is comprised in the area of feminist robotics. We do not propose the feminist perspective in robot design and development as the only viable path to improve robots decisions and mitigate biased outputs. Nevertheless, different feminists theories can provide substantial insights towards a future with more robots coexisting with us. Furthermore, feminist perspectives can guide us in creating robots that not only stop replicating historical discrimination but also impact society to promote and accelerate equality. Previously, cyberfeminism was perceived by Sadie Plant as an ”alliance between women, machinery, and new technology” implying the conception of mechanisms, tools, or strategies towards equality. In this sense, feminist robotics also provides an encouraging field with the potential of structuring a better future. Therefore, we expect more diverse identities to get involved in the development of robots cohabiting with us.

\section{PRECEDENTS OF FEMINISM AS AN AGENDA FOR CHANGE IN AI AND ROBOTICS}
Technologies are created to benefit society and positively change the world. In addition, we expect their advantages to generate better living conditions for us as humans (and hopefully, for all living beings that share this planet with us). Likewise, technology should not at any time harm or unequally benefit people. Technologies must be designed to have an explicitly positive and uniform performance across the diversity of their users. Nevertheless, in recent years, it has been demonstrated that the effects of technological advances are not uniform. Particularly, it was shown that algorithms trained with biased data generate biased outputs \cite{bolukbasi2016man, caliskan2017semantics, buolamwini2018gender, fuchs2018dangers}. What is more, it has been pointed out that discrimination in technological developments also arises from the lack of diverse identities involved in the development process \cite{birhane2019algorithmic}, set up of sensors \cite{wilson2019predictive}, models definition, and evaluation procedure \cite{hurtado2021learning}. Different works exploring fairness can be found in areas such as natural language processing \cite{bolukbasi2016man,lu2020gender}, computer vision \cite{buolamwini2018gender}, robotics \cite{hurtado2021learning}, among others. These studies and daily life experiences confirm that technological developments can present a noticeable disadvantage for less represented identities such as women, sexual-orientation minorities, people of color, people of lower socioeconomic status, and people with disabilities \cite{west2020redistribution,garcia2016racist}. With the potential and dangerous risk of replicating and amplifying historical discrimination in AI and robotics, it is not exceptional that different scholars present feminist perspectives to criticize and propose structural, social, and technological changes in recent decades.

In the multidisciplinary intersection of technology and feminism, we find Haraway’s cyborg manifesto. Here, Haraway provides a feminist insight by means of the fictional figure of a cyborg as the combination of animal and machine, encouraging to overcome the limitations of the traditional roles of gender \cite{donna1991cyborg}. Moreover, Suchman explores a set of feminist principles to consider when designing technologies  \cite{suchman2002located}, while Noble presents the term algorithmic oppression to define the negative effects of algorithms on specific groups of people \cite{noble2018algorithms}. Based on black-intersectional feminism, she studies how the output of algorithms can disproportionately affect people depending on their gender and race. Recently, the Feminist Data Manifest-No presents a joint statement rejecting the use of harmful data and engaging in creating new data available to train fairer models \cite{cifor2019feminist}. In a similar direction, the book Data Feminism is presented to point out data ethics considering ideas of intersectional feminism \cite{d2020data}. More closely, in the Human-Robot Interaction (HRI) field, feminist considerations have been proposed in the design of robots \cite{bardzell2010feminist,winkle2021boosting}. Additionally, numerous works address questions regarding the genre-related attributes of robots and how robots are perceived according to those attributes \cite{nomura2017robots,esposito2020human}. All these guidelines and discussions are important towards the development of more equitative technology.

How can we see this approach in technology, most directly in learned models for robot decision-making? Let us think more directly about the perspective of feminisms in the aforementioned cases. This provides a holistic and integral perspective to problematic situations.  Moreover, it not only demonstrates the importance and scope of several critical perspectives, movements and organizations in the world, but in the same sense, it also evidences the need for an enhancement in technologies that are increasingly present in our lives. Consequently, the process in which robots learn to perform tasks and the effects on society must be constantly analyzed. For whom are robots created, who are they benefiting and who is involved in their development? What learning techniques are being used?

\section{LEARNING PROCESS FOR ROBOTS TO ACQUIRE INTERACTION ABILITIES}

Currently, learning-based approaches play an essential role in the development of robots able to perform complex tasks. These learning approaches are composed of many stages to define the general procedure, data to use, task to learn, training setup, testing environment, and evaluation metrics, to name a few. From the technical point of view, Supervised Learning (SL) and Inverse Reinforcement Learning (IRL) are two techniques that have gained particular attention in the past years in the robot learning field. SL uses data and annotations as learning guidance. The data is usually gathered from the real world or simulations. In this case, SL uses both data and annotations during the learning process to optimize a model with the aim of generating predictions that are as similar as possible to the wanted output (annotations). Therefore, SL requires a careful annotation process that relates the input data and the correct result. On the other hand, IRL is a technique that allows the recovery of a cost function that explains an observed behavior. In this case, the model aims to imitate an "expert" behavior. This allows the model learning interaction abilities that clone the demonstrations of the expert.

It is clear the human involvement in the different stages of robot development. From the social and technical point of view, human decisions directly influence the final model with which the robot is going to operate. It has been studied before how multiple social and technical factors can lead to bias while learning navigation models \cite{hurtado2021learning}. The reported factors include data-related issues, bias induced in the model optimization, evaluation-induced bias, and the complexity of defining a fairness metric. We explore in more detail these factors in the following section.


\section{THE IMPACT OF THE MODELS LEARNED BY THE ROBOT}
In the past years, different situations have been highlighted in which AI algorithms and robots have explicit discrimination outcomes. Such is the case of RGB-based detection systems that disproportionately fail on pedestrians of darker skin color \cite{wilson2019predictive}, the race bias in the facial recognition applications \cite{buolamwini2018gender}, and the gender bias found in the hiring system of amazon \cite{kodiyan2019overview}. However, other less explicit discrimination could be incorporated, especially in robots operating around humans. These situations have been explored in the learning algorithms that are fed with data that include systematic discrimination, when discrimination factors are included through the learned model or biased test methodologies are unable to properly detect the unwanted results \cite{hurtado2021learning}. In other words, bias in robotics is a problem with great potential to amplify and perpetuate problems in society, and although bias in terms of race and gender is more evident, it is not the only bias that can be incorporated into robots.

In the specific case of robotics, critical analysis from different social perspectives is important since many discriminations and injustices happen on the physical spectrum. Unlike algorithms, robots operate and learn from physical interactions. For this reason, it is crucial to question the processes of creating a robot that will operate around humans. Not only because of the risk of physical harm to people but also because of the complexity of the learning models obtained that are susceptible to complex social problems. Although algorithms can learn discrimination through historical data, some discriminations can be included during the learning process, especially the discrimination linked to the behavior and reactions we have as people in social situations.

The relevance of analyzing from different perspectives robots with the potential of influencing the lives of diverse people is exposing current social problems that are at risk of being incorporated in the robot's model. In this case, understanding the factors of the learning process that are susceptible to replicate social bias is important to prevent unwanted outcomes that perpetuate structures of discrimination. Therefore, taking a step to generate tools that allow maximizing the benefit of robots. Consequently, instead of robot behavior reflecting problems in society, we can promote better dynamics and relationships.

\section{FEMINIST PERSPECTIVE ON ROBOT LEARNING PROCESS}
From feminisms, the ability to understand from the intersectional lens the established dynamics has allowed us to diversify and reconfigure numerous spaces. We understand that the goal of a critical perspective is not the recognition of identities. We understand that the identification of these has fallen into co-optation of discourses at the service of the market. The primary purpose of this paper is to think of a world where people can abolish traditional social categories that have discriminated against diverse people across history. Robotics is an appropriate field to evaluate social categories,
roles, and dynamics if we develop robots as social agents and provide them with social skills for human safety and comfort.

Within feminisms, the theoretical stream has developed several concepts to explain how oppressions shape human relations and consequently people. One such concept is intersectionality. This approach recognizes the interconnectedness of different categories of inequality: gender, ethnicity and class, age. Additionally, it incorporates the concept of subjectivities and experiences of bodies situated in diverse historical contexts. It was firstly introduced by Kimberle Crenshaw with the conceptualization of  black feminism and its political view of how antiracist politics are integral, since racism is a structure, one that cements society \cite{crenshaw1989demarginalizing}. This approach follows the same critique developed by feminists against single essentialist thinking, and it emphasizes the importance of diverse and multicultural historical contexts.

Decolonial feminism states that oppressions are not a hierarchy but an assemblage that privileges whiteness and the global north. For Viveros, social relations are consubstantial and co-extensive \cite{vigoya2016interseccionalidad}. They are consubstantial inasmuch as they generate experiences that cannot be divided sequentially but for analytical purposes, and they are co-extensive because they co-produce each other  \cite{vigoya2016interseccionalidad}. Quijano \cite{quijano2000coloniality}, one of the precursors of decolonial theory in Latin America explains how colonialism is institutionalized through knowledge as he highlights how sciences are measured by the global north structure. Paulo Freire wrote “Pedagogy of the oppressed”, in which he longs for an emancipatory education, one where the liberation from structures can be thought of. He reflects on the relationship between student and teacher and how in order to think of an integral process, these two must be at the same level \cite{freire2018pedagogy}. Therefore, allowing them to learn from each other, dismantling the hierarchies that are imposed on us from the institutional framework. Here, Freire argues that the deconstruction and subsequent equitable relationship is the path of an emancipatory guidance \cite{freire2018pedagogy}.

In community feminism, the theoretical basis is to think of the community as a horizontal assembly of relationships that understand the colonial heritage that tries to classify them. Community feminist thinkers such as Adriana Guzma´n state that just as we learn in our communities that are mediated by racism, classism and sexism to relate to each other we relate to nature. Community feminism is a social theory that can discuss with other theories to raise tools and proposals.

The aforementioned Global South theories, allow us to guide the initial discussion of the complex process of thinking of an integral learning process for robot decision-making within a feminist perspective. The learning techniques we discuss in this work are often inspired or compared to hegemonic learning processes in humans: such as traditional, technological and conductive. In this frame of learning, different agents have different importance. Differently, in another approach, students are not passive agents, and can have equally importance in the process as the teacher, as remarked by the South Global work in \cite{vigoya2016interseccionalidad}. Learning is an ongoing process that includes continuous research, adaptive pedagogic methods, evaluation of the effectiveness in the current social environment, and analysis of context and data. This is especially important in robot learning that is a complex process with high involvement of human judgment.Taking that into account, how do we analyze the robot learning techniques and information used? How can we make sure it is, in fact, an integral process? Robot learning, and especially in social environments, demand research and advanced tools to equip them to confront complex situations and decisions, just as Humans face them on a daily basis.

From a social perspective, groups of people lacking in diversity have a limited understanding of the outcomes that can cause discrimination as a result of the robot's actions, behavior, and interactions. We have seen similar results from learning models used in technologies deployed in daily environment such as voice recognition that initially struggled to respond to women’s voice commands \cite{bajorek2019voice}. In this case, the women’s interaction with this technology presented a foreseeable systematic failure. From a technical perspective, the human judgment to decide on the different stages of the learning process as well as the society representation in data impact the trained model. It is expected that while using biased data we obtain biased outcomes. A less explored learning stage that can lead to discrimination is the evaluation procedure \cite{hurtado2021learning}. As exposed in \cite{hurtado2021learning}, such is the case of the testing experiments for learned models in robot navigation where no woman was included \cite{kivrak2021social}. If a navigation model is trained and validated for safety and comfort while excluding certain groups of people, we can foresee biased human-robot interaction. 

To make the robots’ decisions integral and not biased by our own historical structures, interculturality and decoloniality offers a perspective for change. This perspective encourages a strength initiative and ethical-moral agency to question and disrupt the established \cite{walsh2010interculturalidad}. With this perspective, it is possible to question the current configuration of the methodologies, creation of work teams, processes to record data, among others. It is important to highlight that using a reduced sample of the population in the process of learning social parameters in robotics is problematic, specially with socio-political tensions explained in the decolonial theory. Similarly, having non-diverse identities in the development team can lead to overall biased judgment. Humanity is rich in diversity and social interactions are extremely complex. In this sense, this work encourages nourishing ourselves with research, proposals, and data from diverse parts of the globe, going outside of the north perspective. As Freire exposes in the student-teacher dynamic, robot learning can be confirmed through integral mutual growth, allowing robots to adapt to novel environments and constantly learn from experiences. Consequently, a feminist perspective in the learning process of robot decision-making, emphasizes the importance of diverse and multicultural historical contexts by remarking the learning process includes unlearning and questioning the predefined standards in robot learning.

\section{CONCLUSIONS}

In this paper, we explored with a feminist perspective possible discriminating behaviors that can be incorporated into robots in the learning process. Besides questioning how these behaviors are represented in learned models to make decisions, we explore why robots can learn biased behavior. It is not unexpected how biased outcomes mainly affect certain communities that have been historically oppressed. In this paper, we used theories that explain how susceptible is the learning process to induce biased results while understanding different historical contexts. We highlighted how robots operate employing a learning process comparable to some extent with humans. Moreover, we acknowledge the influence of human judgment in the stages of robot learning processes. We recognized robotics as a field with potential influence on human relationships.  With our initial discussion, we encourage the conscious development of robots that influence acceptable behaviors by researching and nourishing novel methodologies for robot decision-making. Particularly, feminisms from the global south give us tools to discuss towards a robotic field in which interdisciplinary teams can respond to safety and comfort requirements for robots to operate also under parameters for diversity and inclusion.

\addtolength{\textheight}{-12cm}   





\bibliographystyle{unsrt}
\bibliography{IEEEexample}



\end{document}